\documentclass{article}
\usepackage{spconf,amsmath,graphicx,hyperref}
\usepackage{adjustbox}
\usepackage{colortbl}
\usepackage{multirow}
\usepackage{xcolor}
\usepackage{amssymb}
\usepackage{dashrule}
\usepackage{xspace}

\def\L{{\cal L}}

\title{Domain-Invariant Mixed-Domain Semi-Supervised Medical Image Segmentation with Clustered Maximum Mean Discrepancy Alignment}
\name{%
  \begin{tabular}[t]{@{}c@{}} 
    Ba-Thinh Lam$^5$ \qquad Thanh-Huy Nguyen$^1$ \qquad  Hoang-Thien Nguyen$^{1,2}$ \qquad Quang-Khai Bui-Tran$^{2,3}$ 
    \\ 
    Nguyen Lan Vi Vu$^1$ \qquad Phat K. Huynh$^2$
     \qquad Ulas Bagci$^4$ \qquad Min Xu$^1$* \thanks{* Corresponding Author: mxu1@cs.cmu.edu}
  \end{tabular}%
}

\address{
$^1$Carnegie Mellon University, PA, USA \\
$^2$PASSIO Lab, North Carolina A\&T State University, NC, USA \\
$^3$Ho Chi Minh University of Science, Vietnam \\
$^4$Northwestern University, IL, USA\\
$^5$University of North Carolina at Charlotte, NC, USA
}
\newcommand{\cpm}{CPM\xspace}
\newcommand{\modulecluster}{Clustering Module\xspace}
\newcommand{\moduledomain}{MMD-based Domain-Aware Module\xspace}
\newcommand{\cmmd}{CMMD\xspace}
\everypar{\looseness=-1}
\begin{document}
%
\maketitle
\begin{abstract}
Deep learning has shown remarkable progress in medical image semantic segmentation, yet its success heavily depends on large-scale expert annotations and consistent data distributions. In practice, annotations are scarce and images are collected from multiple scanners or centers, leading to mixed-domain settings with unknown domain labels and severe domain gaps. Existing semi-supervised or domain adaptation approaches typically assume either a single domain shift or access to explicit domain indices, which rarely hold in real-world deployment. In this paper, we propose a domain-invariant mixed-domain semi-supervised segmentation framework that jointly enhances data diversity and mitigates domain bias. A \textbf{Copy-Paste Mechanism (CPM)} augments the training set by transferring informative regions across domains, while a \textbf{Cluster Maximum Mean Discrepancy (CMMD)} block clusters unlabeled features and aligns them with labeled anchors via an MMD objective, encouraging domain-invariant representations. Integrated within a teacher–student framework, our method achieves robust and precise segmentation even with very few labeled examples and multiple unknown domain discrepancies. Experiments on Fundus and M\&Ms benchmarks demonstrate that our approach consistently surpasses semi-supervised and domain adaptation methods, establishing a potential solution for mixed-domain semi-supervised medical image segmentation.
\end{abstract}



%
\begin{keywords}
Semi-supervised learning, Domain Adaptation, Medical Image Segmentation
\end{keywords}
\section{Introduction}
\label{sec:intro}

\begin{figure}[!ht]
    \centering
    \includegraphics[width=\linewidth]{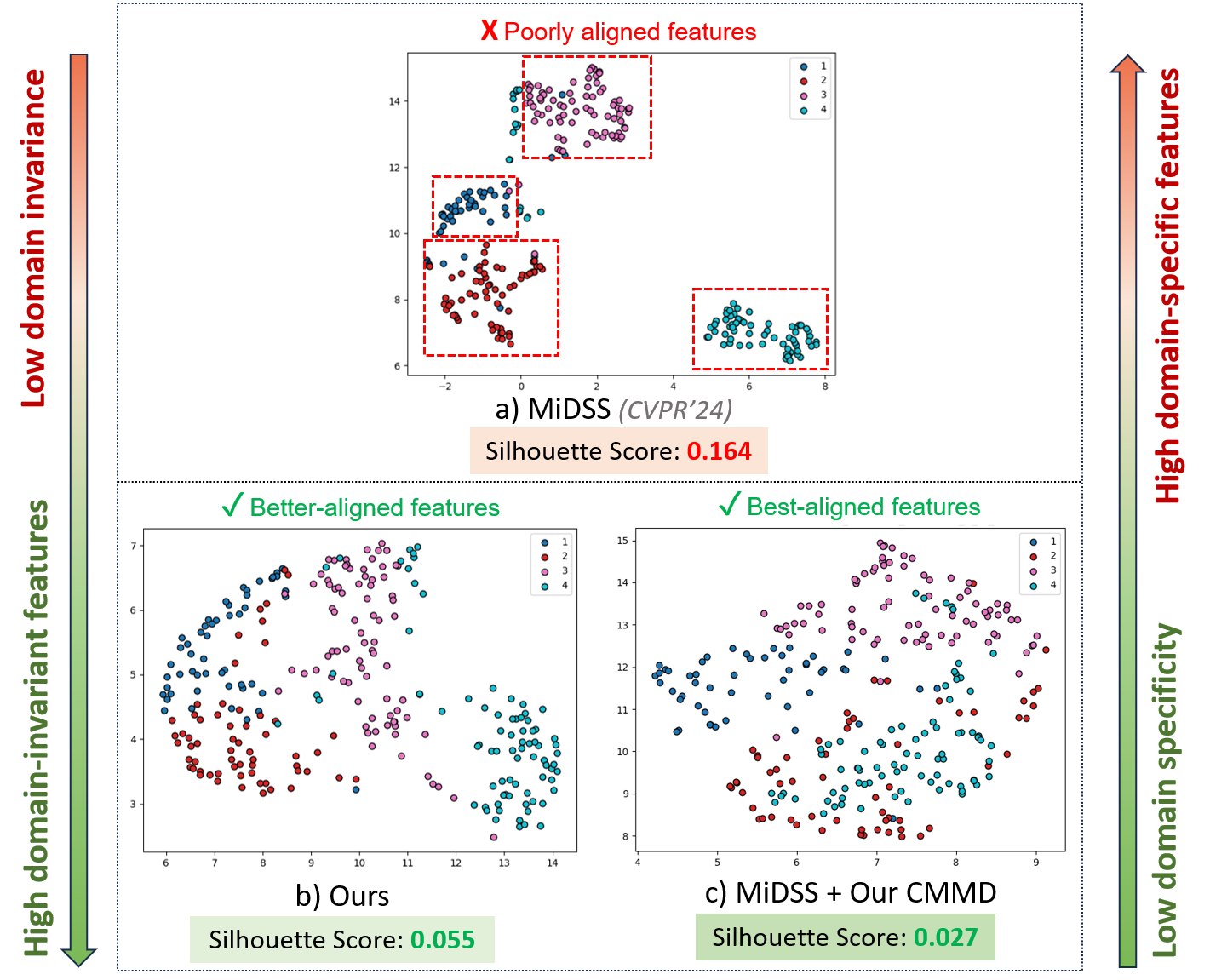}
    \caption{Feature visualizations of the student encoder’s last layer on the Fundus dataset using UMAP. Domains are shown in different colors, and the model was trained with labeled samples from Domain 4. Silhouette Score assesses cluster quality, with higher scores indicating more separated clusters.}
    \label{fig:feature_visualization}
\end{figure}

Medical image segmentation is a potential task for computer-aided diagnosis, yet building reliable models often requires a large quantity of expert annotations. In practice, collecting such labels is expensive and time-consuming, and the problem becomes even more challenging when data originate from multiple sources, scanners, or clinical centers. 
Models trained with a few labeled examples tend to overfit, while standard semi-supervised learning (SSL) \cite{coranet, sohn2020fixmatch, bai2023bidirectional,yu2019uncertainty, chen2021semi, FDCL, caussl}
or domain adaptation (DA) \cite{survey_da, FDA, udavae, tzeng2017adversarial, mme} approaches frequently assume either a single target distribution or access to explicit domain labels, which rarely hold in real-world deployments.

Recent studies have begun to address this setting, often referred to as \emph{mixed-domain semi-supervised segmentation}. 
Methods such as MiDSS~\cite{ma2024constructing} construct intermediate domains to bridge distribution gaps, showing encouraging results on benchmarks like Fundus \cite{fundus} and M\&Ms \cite{mms}. However, this approach still does not solve the domain discrepancies thoroughly. Indeed, we investigate this by visualizing feature representations from pretrained models on Fundus. Fig. \ref{fig:feature_visualization}a shows learned features from MiDSS \cite{ma2024constructing} cluster more separated by domains with a high Silhouette Score. This indicates that MiDSS produces domain-specific features instead of domain-invariant features. Plus, we can not directly apply common DA approaches to solve, due to the unknown domains and unprovided domain labels.

To address these challenges, we propose a simple yet effective framework that simultaneously enhances data diversity and mitigates domain bias explicitly in input and feature spaces. Specifically, we design a \textbf{Copy-Paste Mechanism (CPM)} to augment the training set by transferring informative regions across domains, thereby expanding the sample space under limited annotations. Also, a \textbf{Cluster Maximum Mean Discrepancy (CMMD)} block is introduced to cluster unlabeled features and align them with labeled anchors through an MMD loss, promoting domain-invariant representations. By combining these modules within a teacher–student network, our method explicitly addresses multiple mixed-domain discrepancies and produces precise segmentation.

\label{sec:method}

\begin{figure*}[h]
    \centering
    \includegraphics[width=\linewidth]{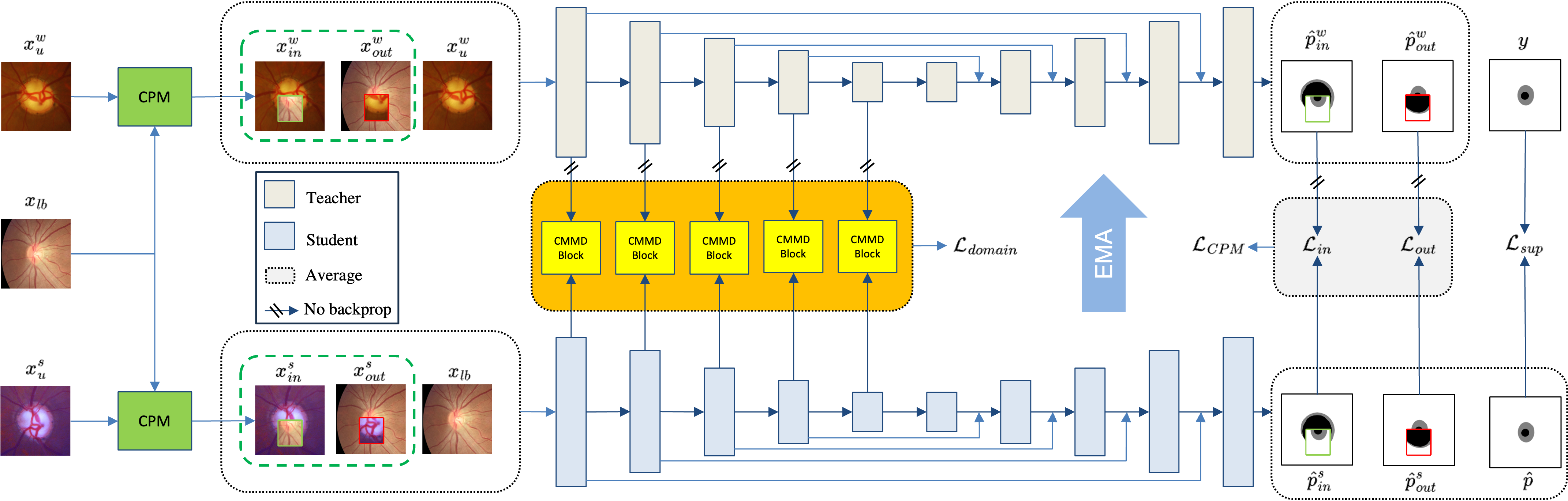}
    \caption{Overall illustration of Our Framework.}
    \label{fig:framework}
\end{figure*}


\section{Methodology}
\subsection{Problem Definition}
\label{subsec:problem}

Let $\mathcal{S}_l = \{(x_i, y_i)\}_{i=1}^N$ be a small labeled set from domain $D_j$, and $\mathcal{S}_u = \{x_i\}_{i=1}^M$ a large unlabeled set from a mixture of domains $\{D_1, \dots, D_K\}$, with $K$ unknown.  
The aim is to learn a segmentation model $f_\theta$ that generalizes across all domains using few labeled and many unlabeled samples.

This scenario differs from classical DA or SSL in two ways: (1) the unlabeled pool spans several hidden domains with no domain labels, and (2) $K$ is not fixed or known, making standard DA inapplicable. Bridging $D_j$ and $\{D_k\}$ thus requires discovering domain structure and mitigating discrepancies while preserving semantics.

As shown in Fig.~\ref{fig:framework}, our framework tackles this by (i) enlarging the cross-domain training space via CPM, and (ii) learning domain-invariant features with CMMD, which clusters unlabeled features and aligns them to labeled anchors through an MMD objective.

\subsection{Copy-Paste Mechanism}
To enrich the sample for model learning, we combine weak- and strong-augmented unlabeled examples $x^{w}_u,x^{s}_u$ with labeled examples $x_{lb}$ through Copy-Paste Mechanism (CPM):
\begin{align}
    x^w_{in} &= (1-\mathcal{M}) \odot x^w_{u} + \mathcal{M} \odot x_{lb},\\
    x^w_{out} &= \mathcal{M} \odot x^w_{u} + (1-\mathcal{M}) \odot x_{lb},\\
    x^s_{in} &= (1-\mathcal{M}) \odot x^s_{u} + \mathcal{M} \odot x_{lb},\\
    x^s_{out} &= \mathcal{M} \odot x^s_{u} + (1-\mathcal{M}) \odot x_{lb},
\end{align}
where $\mathcal{M}$ is a binary mask ($1$ indicates the region of interest). The models generate predictions for these mixed examples:
\begin{align}
    &\hat{p}^w_{in} = T(x^w_{in};\theta^T), 
    &\hat{p}^s_{in} = S(x^s_{in};\theta^S)\\
    &\hat{p}^w_{out} = T(x^w_{out};\theta^T), &\hat{p}^s_{out} = S(x^s_{out};\theta^S)
\end{align}
we use $\hat{p}^w_{in}$ and $\hat{p}^w_{out}$ as pseudo-labels to guide $\hat{p}^s_{in}$ and $\hat{p}^s_{out}$ by consistency losses $\L_{in}$ and $\L_{out}$:
\begin{align}
    \L_{in} &= Dice(\hat{p}^s_{in}, \hat{p}^w_{in}) + CE(\hat{p}^s_{in}, \hat{p}^w_{in}), \\
    \L_{out} &= Dice(\hat{p}^s_{out}, \hat{p}^w_{out}) + CE(\hat{p}^s_{out}, \hat{p}^w_{out}).
\end{align}
Finally, the \cpm loss is computed as follows:
\begin{align}
    \L_{CPM} &= \L_{in} + \L_{out}.
\end{align}

\subsection{Cluster Maximum Mean Discrepancy Block}
\begin{figure}[!ht]
    \centering
    \includegraphics[width=\columnwidth]{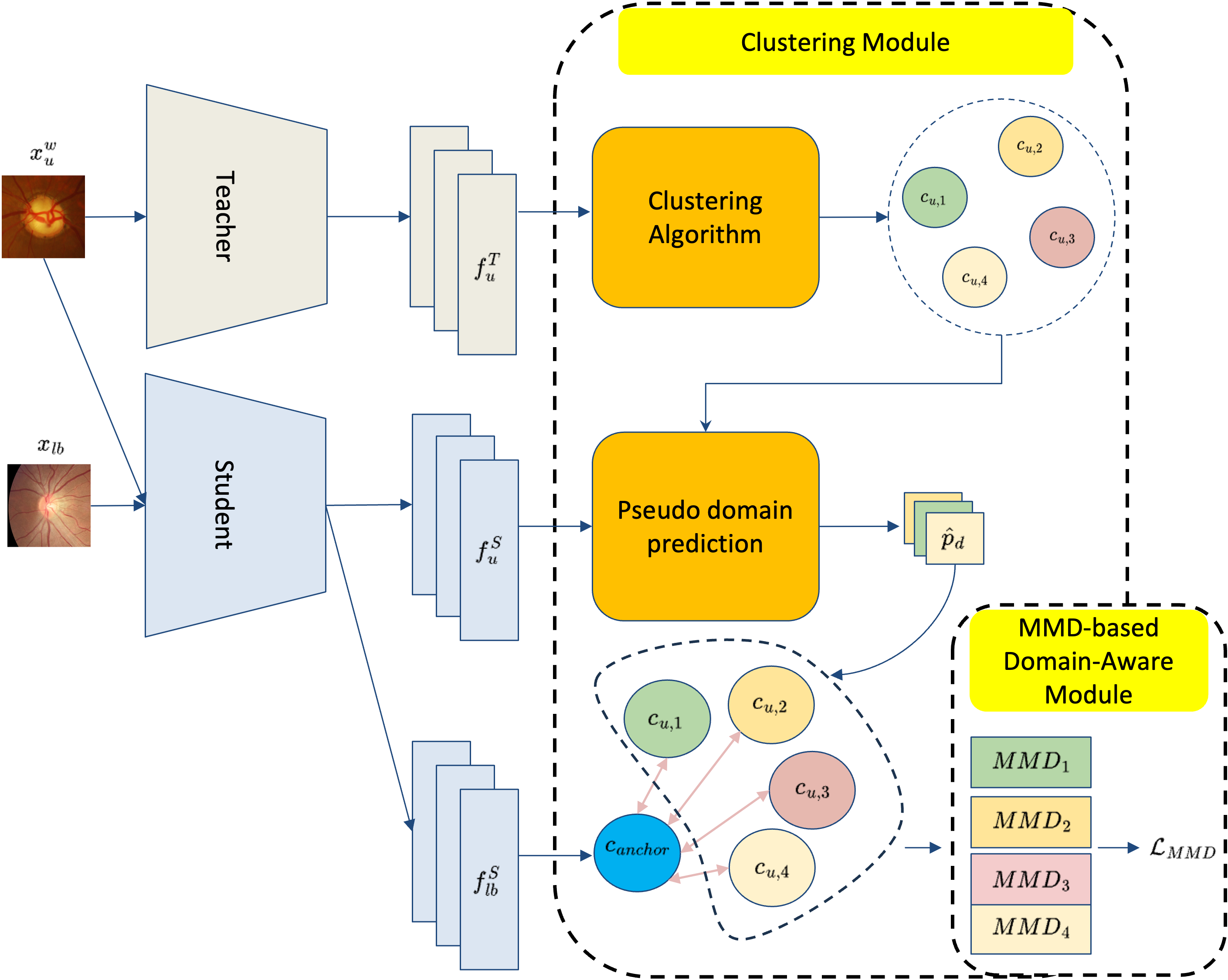}
    \caption{Illustration of CMMD Block.}
    \label{fig:cmmd}
\end{figure}
To handle inter-domain discrepancies, we propose the Cluster Maximum Mean Discrepancy Block (CMMD), composed of a \modulecluster and a \moduledomain (Fig.~\ref{fig:cmmd}). The \modulecluster estimates clusters from feature-space density, while the \moduledomain pulls them together by minimizing their Maximum Mean Discrepancy. CMMD is applied to all encoder layers, ensuring domain signals are removed before final decoder predictions.

\subsubsection{Clustering Module}
Given the teacher model's stability and generalization, we cluster unlabeled features $x_u$ extracted from it. The teacher first processes weakly augmented inputs $x_u^w$ to obtain features $f^T_u$ at a chosen encoder layer:
\begin{equation}
   f^T_u = E^T(x_u^w).
\end{equation}

Because the number of domains is unknown, clustering is challenging. We address this by estimating clusters from feature-point density using Hierarchical Density-Based Spatial Clustering of Applications with Noise (HDBSCAN \cite{hdbscan}):

\begin{equation}
\begin{split}
  (c_{u,1}, \ldots, c_{u,Q}) 
    = \mathrm{HDBSCAN}\big(f_u^w, \textit{minClusterSize}, \textit{minSample}\big),
\end{split}
\end{equation}
where $c_{u,i}$ is the centroid of cluster $C_{u,i}$ for unlabeled features, and $Q$ is the predicted number of domains. As they update over epochs, these centroids are termed \textit{pseudo-centroids}.

They are then used to assign pseudo-domain labels to weakly augmented features $f_u^S$ extracted by the student encoder $E^S(\cdot; \theta_E^S)$:
\begin{equation}
   f_u^S = E^S(x_u^w).
\end{equation}

Distances to each pseudo-centroid are computed, and the label $\hat{p}_d$ is set to the closest cluster:
\begin{align}
    d_i &= \| f_u^S - c_{u,i} \|_2 , \quad 1 \leq i \leq Q,\\
    \hat{p}_d &= \arg\min_{1 \leq i \leq Q} d_i.
\end{align}

\subsubsection{MMD-based Domain-Aware Module}
To reduce inter-domain gaps, we minimize Maximum Mean Discrepancy (MMD) between clusters. Computing MMD for all pairs is costly, so we pick a reliable anchor cluster and compute MMD w.r.t. it, pulling clusters together efficiently. Labeled features $f_{lb}^S$ act as the anchor $C_{anchor}$, since they are drawn i.i.d. from domain $D_j$. For cluster $C_{u,i}$:
\begin{equation} 
\begin{split} MMD_i &= \mathrm{MMD}^2(C_{u,i}, C_{\text{anchor}}) \\ &= \frac{1}{m_i(m_i-1)} \sum_p \sum_{q \neq p} k(\mathbf{c}^{(u,i)}_p, \mathbf{c}^{(u,i)}_q) \\ &\quad - \frac{2}{m_in} \sum_p \sum_q k(\mathbf{c}^{(u,i)}_p, \mathbf{c}^{(\text{anchor})}_q) \\ &\quad + \frac{1}{n(n-1)} \sum_p \sum_{q \neq p} k(\mathbf{c}^{(\text{anchor})}_p, \mathbf{c}^{(\text{anchor})}_q), 
\end{split} 
\end{equation} where $k(\mathbf{x}_i, \mathbf{x}_j) = \exp\left( -\frac{\|\mathbf{x}_i - \mathbf{x}_j\|^2}{2\sigma^2} \right)$ is the gaussian kernel, $m_i$ is the number of features belonging to $C_{u, i}$, and $n$ is the number of features belonging to $C_{anchor}$. These MMD values are averaged to compute MMD loss $\L_{MMD}$, which is minimized to address the domain gaps: 
\begin{equation} 
\L_{MMD} = \frac{1}Q{}\sum_{i=1}^{Q} MMD_i. 
\end{equation}

\subsubsection{Domain Loss}
While \cmmd can handle domain gaps , the features extracted at layers of the decoder still remain domain signal due to skip connections among layers from the encoder and decoder. Plus, applying directly \cmmd for layers of the decoder may harm class predictions of the model, because these decoder features bring more information about class decisions than the domain information. 

To address this issue, we apply \cmmd for all layers of the encoder to remove domain information from intermediate features of the encoder before transferring to the decoder through skip connections. In this way, we compute MMD loss $\L_{MMD}^l$ at the $lth$ layer, respectively. Eventually, the Domain loss $\L_{domain}$ is computed as follows: 
\begin{equation}
    \L_{domain} = \frac{1}{L}\sum_{l=1}^{L}\L_{MMD}^l,
\end{equation}
where $L$ is the number of encoder layers ($L=5$, commonly).

\subsection{Final Loss}
The Domain loss is combined with the CPM loss and the Supervised loss to compute the total loss:
\begin{equation}
    \L_{total} = \L_{sup} + w (\L_{CPM} + \L_{domain}),
\end{equation}
where $w$ is a dynamic coefficient updated by an iteration-dependent Gaussian function, $w=\exp(-5(1-\frac{iter}{iter_{max}}))$.
\section{Experimental Results}
\label{sec:experiments}

\subsection{Datasets \& Metrics}
We evaluate on Fundus and M\&Ms using Dice (DC), Jaccard (JC), Hausdorff distance (HD), and Average Surface Distance (ASD). Both teacher and student adopt UNet~\cite{ronneberger2015u}. Training uses $30$k and $60$k iterations for Fundus and M\&Ms, respectively. HDBSCAN is set with \textit{minClusterSize}=10 and \textit{minSample}=5. All runs are on A5000/RTX3090 GPUs.

\subsection{Results}
For fair comparison, we reproduced the results of MiDSS \cite{ma2024constructing} for all experiments.

\looseness=-1
\textbf{Fundus.} With only 20 labeled images (Table~\ref{tab:fundus}), SSL baselines (UA-MT~\cite{yu2019uncertainty}, CPS~\cite{chen2021semi}) show modest gains, while FixMatch~\cite{sohn2020fixmatch} and BCP~\cite{bai2023bidirectional} remain vulnerable to hidden domains. MiDSS mitigates but does not eliminate domain gaps. Our method, combining \cpm and \cmmd, enlarges the training space via mixed-domain patches and aligns features through an MMD loss, achieving a steady edge over MiDSS (DC $\approx87\%$).
\looseness=-1

\begin{table}[htp]
\centering
\caption{Comparison of different methods on Fundus dataset. The number of labeled examples is 20. Higher is better for DC/JC, lower is better for HD/ASD. Best in \textbf{bold}, second-best \underline{underlined}.}
\label{tab:fundus}
\resizebox{\columnwidth}{!}{
\begin{tabular}{l|cccc|cccc}
\hline
\multirow{2}{*}{Method} & \multicolumn{4}{c|}{DC $\uparrow$ (Optic Cup / Disc)} & Avg. & Avg. & Avg. & Avg. \\
\cline{2-5}
 & Domain 1 & Domain 2 & Domain 3 & Domain 4 & DC $\uparrow$ & JC $\uparrow$ & HD $\downarrow$ & ASD $\downarrow$ \\
\hline
U-Net & 59.54 / 73.89 & 71.28 / 74.23 & 50.87 / 64.29 & 35.61 / 63.30 & 61.63 & 52.65 & 48.28 & 28.86 \\
UA-MT  & 59.35 / 78.46 & 63.08 / 74.45 & 35.24 / 47.73 & 36.18 / 55.43 & 56.24 & 47.00 & 48.64 & 31.35 \\
FDA  & 76.99 / 89.94 & 77.69 / 89.63 & 78.27 / 90.96 & 64.52 / 74.29 & 80.29 & 71.05 & 16.23 & 8.44 \\
FixMatch  & 81.18 / 91.29 & 72.04 / 87.60 & 80.41 / 92.95 & 74.58 / 87.07 & 83.39 & 73.48 & 11.77 & 5.60 \\
CPS  & 64.53 / 86.25 & 70.26 / 86.97 & 42.92 / 54.94 & 36.98 / 46.70 & 61.19 & 52.69 & 34.44 & 26.79 \\
CoraNet  & 61.64 / 87.32 & 65.56 / 87.05 & 66.12 / 83.54 & 49.01 / 77.73 & 72.25 & 60.50 & 20.52 & 10.44 \\
UDA-VAE++  & 55.01 / 80.76 & 68.87 / 85.94 & 63.23 / 84.92 & 68.42 / 80.89 & 73.51 & 61.40 & 17.60 & 9.86 \\
BCP & 71.65 / 91.10 & 77.19 / 92.00 & 72.63 / 90.77 & 77.67 / 91.42 & 83.05 & 73.66 & 11.05 & 5.80 \\
CauSSL & 63.38 / 80.60 & 67.52 / 80.72 & 49.53 / 63.88 & 39.43 / 49.43 & 61.81 & 51.80 & 41.25 & 23.94 \\
MiDSS (reproduced)  & 82.17 / 95.35 & 77.56 / 90.55 & 82.23 / 92.89 & 83.34 / 91.22 & \underline{86.91} & \underline{78.12} & \textbf{8.24} & \textbf{3.95} \\
\hline
Ours & 81.23 / 94.84 & 78.39 / 89.26 & 83.37 / 92.84 & 84.09 / 91.92 & \textbf{86.99} & \textbf{78.22} & \underline{8.25} & \underline{3.96} \\
\hline
\end{tabular}}
\end{table}
\textbf{M\&Ms.}  
Table~\ref{tab:mmms} reports results on M\&Ms with 20 labeled images. FixMatch and BCP perform well but leave scanner gaps. MiDSS improves via intermediate domains yet retains bias. Combining \cpm and \cmmd yields the best scores (DC $\approx84.4$, JC $\approx75$).


\begin{table}[t]
\centering
\caption{Comparison of different methods on M\&Ms dataset. The number of labeled examples is 20. Higher is better for DC/JC, lower is better for HD/ASD. Best in \textbf{bold}, second-best \underline{underlined}.}
\label{tab:mmms}
\resizebox{\columnwidth}{!}{
\begin{tabular}{l|cccc|cccc}
\hline
\multirow{2}{*}{Method} & \multicolumn{4}{c|}{DC $\uparrow$ (LV / MYO / RV)} & Avg. & Avg. & Avg. & Avg. \\
\cline{2-5}
 & Vendor A & Vendor B & Vendor C & Vendor D & DC $\uparrow$ & JC $\uparrow$ & HD $\downarrow$ & ASD $\downarrow$ \\
\hline
U-Net & 57.29 / 37.85 / 34.65 & 73.44 / 64.20 / 53.58 & 55.83 / 48.47 / 44.84 & 63.85 / 52.25 / 49.85 & 53.01 & 44.30 & 38.07 & 22.88 \\
UA-MT  & 38.02 / 25.51 / 14.94 & 61.85 / 54.27 / 47.33 & 43.13 / 35.66 / 28.54 & 41.89 / 38.25 / 26.11 & 37.96 & 29.14 & 72.35 & 40.84 \\
FDA  & 61.66 / 36.32 / 34.71 & 80.67 / 70.99 / 56.75 & 73.80 / 63.62 / 58.36 & 77.23 / 68.87 / 64.33 & 62.28 & 53.33 & 25.99 & 16.10 \\
FixMatch  & 87.26 / 77.78 / 77.14 & 91.06 / 82.78 / 79.07 & 87.84 / 80.07 / 78.03 & 90.86 / 81.75 / 81.84 & 82.96 & 73.99 & 6.21 & 3.51 \\
CPS  & 46.40 / 29.01 / 16.70 & 71.48 / 63.08 / 49.39 & 44.38 / 39.43 / 32.42 & 47.71 / 40.75 / 29.75 & 42.54 & 33.82 & 58.30 & 34.94 \\
CoraNet  & 65.70 / 27.79 / 22.16 & 63.32 / 48.63 / 46.56 & 64.89 / 48.59 / 45.30 & 68.38 / 55.88 / 46.79 & 50.33 & 40.54 & 32.98 & 19.22 \\
UDA-VAE++  & 51.14 / 36.20 / 12.99 & 71.95 / 53.16 / 36.68 & 57.88 / 41.64 / 30.19 & 31.71 / 27.32 / 20.48 & 39.28 & 28.82 & 53.90 & 24.94 \\
BCP  & 85.91 / 73.82 / 78.08 & 85.66 / 74.85 / 76.04 & 61.61 / 54.05 / 51.87 & 76.57 / 62.22 / 79.16 & 71.65 & 62.67 & 30.91 & 18.22 \\
CauSSL & 40.20 / 21.93 / 10.46 & 50.99 / 42.66 / 31.94 & 41.05 / 34.00 / 29.95 & 53.78 / 37.92 / 30.44 & 35.44 & 26.73 & 72.90 & 37.99 \\
MiDSS (reproduced) & 89.66 / 80.44 / 80.93 & 89.86 / 84.27 / 85.93 & 89.75 / 79.98 / 76.77 & 89.01 / 81.54 / 83.35 & 84.25 & 75.20 & \textbf{5.08} & \textbf{2.54} \\
\hline
Ours & 90.27 / 80.53 / 80.97 & 90.17 / 84.46 / 85.77 & 90.12 / 79.70 / 75.56 & 89.77 / 81.72 / 83.53 & \textbf{84.35} & \textbf{75.33} & \underline{5.21} & \underline{2.57} \\
\hline
\end{tabular}}
\end{table}

\subsection{Ablation \& Analysis}
We assess our method on Fundus using 20 labeled and the rest unlabeled images.

\textbf{Efficiency of \cpm and \cmmd.}  
Table~\ref{tab:ablation_fundus} reports the effect of each component. The baseline is a UNet teacher–student with cross-entropy and weak-to-strong consistency. Adding \cpm raises DC/JC by $3.83\%/4.88\%$, confirming the benefit of mixed-domain augmentation. Integrating \cmmd further lifts performance to DC $86.99\%$ and JC $78.22\%$, highlighting its ability to reduce domain discrepancies.

\begin{table}[t]
\caption{Ablation study for components of our framework on Fundus dataset.}
\label{tab:ablation_fundus}
\centering
\small
\setlength{\tabcolsep}{3pt}
\renewcommand{\arraystretch}{1.2}
\begin{adjustbox}{width=\columnwidth}
\begin{tabular}{l|c|c|cccc|c|c}
\hline
\multirow{2}{*}{Ablation} & \multicolumn{2}{c|}{Component} & \multicolumn{4}{c|}{DC$\uparrow$} & Avg. & Avg. \\
\cline{2-7}
 & \cpm & \cmmd & Domain 1 & Domain 2 & Domain 3 & Domain 4 & {DC$\uparrow$} & {JC$\uparrow$} \\
\hline
Baseline & & & 74.96 / 93.50 & 71.08 / 85.72 & 77.45 / 89.10 & 81.32 / 91.36 & 83.06 & 73.19 \\
1 & \checkmark & & 82.39 / 95.08 & 77.78 / 90.24 & 82.15 / 92.81 & 84.33 / 90.80 & 86.89 & 78.07 \\
2 & \checkmark & \checkmark & 81.23 / 94.84 & 78.39 / 89.26 & 83.37 / 92.84 & 84.09 / 91.92 & \textbf{86.99} & \textbf{78.22} \\

\hline
\end{tabular}
\end{adjustbox}
\end{table}

\textbf{Plug-in Property of \cmmd.}  
Table~\ref{tab:cmmd_plugin_fundus} shows that adding \cmmd to MiDSS lifts DC/JC to $87.18\%$/$78.45\%$, confirming its ability to reduce domain gaps. Since \cmmd acts in the feature space, it can be easily integrated into other frameworks without extra parameters.

\begin{table}[!t]
\centering
\caption{Comparison of MiDSS before and after integrating our \cmmd on Fundus.}
\label{tab:cmmd_plugin_fundus}
\begin{adjustbox}{width=\columnwidth}
\begin{tabular}{l|cccc|cc}
\hline
\multirow{2}{*}{Method} & \multicolumn{4}{c|}{DC $\uparrow$} & Avg. & Avg. \\
\cline{2-5}
 & Domain 1 & Domain 2 & Domain 3 & Domain 4 & DC $\uparrow$ & JC $\uparrow$ \\
\hline
MiDSS (reproduced) & 82.17 / 95.35 & 77.56 / 90.55 & 82.23 / 92.89 & 83.34 / 91.22 & 86.91 & 78.12 \\
MiDSS + Our \cmmd & 83.21 / 95.63 &	78.36 / 91.48 & 82.02 / 92.71 &	83.30 / 90.75 & \textbf{87.18} & \textbf{78.45}\\
\hline
\end{tabular}
\end{adjustbox}
\end{table}

\textbf{Domain gap alleviation by \cmmd.}  
Fig.~\ref{fig:feature_visualization} shows UMAP embeddings from the 5th encoder layer. MiDSS features form separate clusters (e.g., domain 4 is most isolated), revealing domain gaps. Adding \cmmd yields more mixed representations, indicating reduced domain-specific variation and consistent learning of domain-invariant features.

\section{Conclusion}
\looseness=-1
We introduced a domain-invariant framework for mixed-domain semi-supervised medical image segmentation. By combining the Copy-Paste Mechanism (CPM) with the Cluster Maximum Mean Discrepancy (CMMD) block, our method enriches cross-domain training data and aligns unlabeled features with labeled anchors, effectively mitigating domain gaps explicitly. Integrated into a teacher–student network, it achieves consistent improvements on Fundus and M\&Ms benchmarks, surpassing several semi-supervised learning and domain adaptation methods. Future work will explore extending CMMD to more complicated architectures and additional modalities, further advancing robust segmentation under limited annotations and multiple hidden distribution mismatches.
\looseness=-1

\bibliographystyle{IEEEbib}
\bibliography{refs}

\end{document}